\pdfoutput=1

\documentclass[11pt]{article}
\usepackage{authblk}
\usepackage[]{emnlp2021}
\usepackage{textcomp}
\usepackage{amssymb}
\usepackage{newunicodechar}
\usepackage[mathletters]{ucs}
\usepackage{subcaption}
\usepackage{enumitem}

\usepackage{amssymb}
\usepackage{pifont}
\newcommand{\cmark}{\ding{51}}%
\newcommand{\xmark}{\ding{55}}%

\usepackage{times}
\usepackage{latexsym}

\usepackage[T1]{fontenc}

\usepackage[utf8]{inputenc}

\usepackage{microtype}

\usepackage{multirow}

\usepackage{pgfplots}
\usepackage{siunitx}
\usepackage{xcolor}
\usepackage{tcolorbox}
\usepackage[T1]{fontenc}

\makeatletter
\newcommand*{\affaddr}[1]{#1} 
\newcommand*{\affmark}[1][*]{\textsuperscript{#1}}
\newcommand*{\email}[1]{\texttt{#1}}
\renewcommand\AB@affilsepx{, \protect\Affilfont}
\makeatother

\newcommand\crule[3][black]{\textcolor{#1}{\rule{#2}{#3}}}

\title{GupShup: An Annotated Corpus for Abstractive Summarization of Open-Domain Code-Switched Conversations 
\thanks{Authors 1,2,3, and 4 have equal contributions}}

\author{%
Laiba Mehnaz\affmark[2], Debanjan Mahata\affmark[1], Rakesh Gosangi\affmark[1], Uma Sushmitha Gunturi\affmark[2], \\ Riya Jain\affmark[2], Gauri Gupta\affmark[2], Amardeep Kumar\affmark[2], Isabelle Lee\affmark[3], Anish Acharya\affmark[4], Rajiv Ratn Shah\affmark[2] \\
\small \affaddr{\affmark[1]Bloomberg, USA}, \affaddr{\affmark[2]MIDAS Lab, IIIT-Delhi}, \affaddr{\affmark[3]University of Washington, USA} \small \affaddr{\affmark[4]University of Texas at Austin, USA} \\
\small \email{\{dmahata,rgosangi\}@bloomberg.net}, \email{\{laibamehnaz,gaurigupta.315,adkr6398\}@gmail.com}\\ \small \email{umasgunturi@gmail.com}, \email{riya.j17@iiits.in}, \email{iglee@uw.edu}, \email{anishacharya@utexas.edu} \\ \small \email{rajivratn@iiitd.ac.in}%
}

\begin{document}
\maketitle
\begin{abstract}
Code-switching is the communication phenomenon where speakers switch between different languages during a conversation. With the widespread adoption of conversational agents and chat platforms, code-switching has become an integral part of written conversations in many multi-lingual communities worldwide. This makes it essential to develop techniques for summarizing and understanding these conversations. Towards this objective, we introduce abstractive summarization of Hindi-English code-switched conversations and develop the first code-switched conversation summarization dataset - \textit{GupShup}, which contains over 6,831 conversations in Hindi-English and their corresponding human-annotated summaries in English and Hindi-English. We present a detailed account of the entire data collection and annotation processes. We analyze the dataset using various code-switching statistics. We train state-of-the-art abstractive summarization models and report their performances using both automated metrics and human evaluation. Our results show that multi-lingual mBART and multi-view seq2seq models obtain the best performances on the new dataset\footnote{We are in the process of making the dataset publicly available. Please contact Debanjan Mahata at dmahata@bloomberg.net for getting the latest updates}.

\end{abstract}

\section{Introduction}
\label{sec:intro}

Conversation summarization is the process of generating a condensed version of a given conversation while preserving the most salient aspects. With the widespread use of various chat applications such as messaging apps and virtual assistants  \cite{klopfenstein2017rise}, there has been a growing interest in the abstractive summarization of written conversations \cite{mehdad2014abstractive,goo2018abstractive,zhao2019abstractive}. \\


\begin{table}[h]
    \centering
    \small
    \begin{tabular}{|p{7.5cm}|}
    \hline
\textbf{Leon}: \textcolor{magenta}{kya tujeh abhi tak naukari nahi mili}? \\
\textbf{Arthur}: \textcolor{magenta}{nahi} \textcolor{blue}{bro}, \textcolor{magenta}{abhi bhi} \textcolor{blue}{unemployed} :D \\
\textbf{Leon}: \textcolor{blue}{hahaha, LIVING LIFE} \\
\textbf{Arthur}: \textcolor{magenta}{mujeh yeh bahot acha lagta hai}, \textcolor{magenta}{dopahar ko jagata hoon}, \textcolor{blue}{sports} \textcolor{magenta}{dekhta hoon} - \textcolor{magenta}{ek aadmi ko aur kya chahiye}?\\
\textbf{Leon}: \textcolor{blue}{a paycheck}? ;) \\
\textbf{Arthur}: \textcolor{blue}{mean} \textcolor{magenta}{mat bano} ... \\
\textbf{Leon}: \textcolor{blue}{but seriously}, \textcolor{magenta}{mere dosth ke} \textcolor{blue}{company} \textcolor{magenta}{mein ek} \textcolor{blue}{junior project manager offer} \textcolor{magenta}{hai}, \textcolor{magenta}{tujeh} \textcolor{blue}{interest} \textcolor{magenta}{hai}? \\
\textbf{Arthur}: \textcolor{blue}{sure thing}, \textcolor{magenta}{tere pass} \textcolor{blue}{details} \textcolor{magenta}{hai}? \\ 
\textbf{Leon}: <file\_photo> \\
\hline
\textbf{English Summary}: \textit{Arthur \textcolor{blue}{is still unemployed}. Leon \textcolor{blue}{sends him a job offer for junior project manager position}. Arthur \textcolor{blue}{is interested}.} \\
\hline
\end{tabular}
\caption{An example of a code-switched Hindi-English conversation and the corresponding English summary. Color coding: English words are in \crule[blue]{0.25cm}{0.25cm}, transliterated Hindi words are in \crule[magenta]{0.25cm}{0.25cm}, and language agnostic words such as named entities and punctuations are in (\crule[black]{0.25cm}{0.25cm})}
\label{tab:sample_conversation}
\end{table}

One of the biggest challenges in summarizing written conversations has been the lack of large datasets with human-annotated summaries. Most researchers evaluate their conversation summarization techniques on transcriptions of AMI \cite{carletta2005ami} or ICSI meeting corpus \cite{janin2003icsi} using meeting topics as a proxy for the summary. These corpora are very useful for various speech-related research problems, but they are not representative of written conversations in chat applications. Recently, \cite{gliwa2019samsum} published the SAMSum corpus, which contains over 16,000 written English conversations with their corresponding manually annotated summaries. Though these conversations were not extracted from actual chat applications, they were created by linguists to replicate natural conversations. To our knowledge, this is the largest summarization dataset for written conversations. 

The SAMSum corpus is completely monolingual (English); therefore, any models trained on this dataset may not adapt effectively to multi-lingual or especially code-switched conversations when speakers alternate between different languages within the scope of a conversation or even an utterance \cite{gumperz1977sociolinguistic,muysken2000bilingual,myers1997duelling}. Code-switching is commonly observed during interactions between peers who are fluent in multiple languages. For example, in the Indian subcontinent, it is common for people to alternate between English and other regional languages like Hindi over the course of a single conversation. This behavior is termed as Code-switching. It is an integral part of both written and spoken conversations for various multi-lingual communities across the world \cite{auer2013code}. Developing models that can accurately process code-switched text is essential to the proliferation of NLP technologies to these communities and contributes towards the diversity and inclusivity of language resources \cite{joshi2020state}. However, building such models would require high-quality human-curated datasets.

This paper introduces the new task of abstractive summarization of open-domain code-switched written conversations. Namely, given a multi-party conversation in Hindi-English on any topic, the objective is to generate a summary in English, as shown in Table \ref{tab:sample_conversation}. These English summaries can serve as input to other downstream NLP models, which are often trained only on English data, to perform various other tasks such as intent classification, question answering, item recommendation. This task also introduces some exciting challenges to NLP researchers because they need to build models that translate as well as summarize a given conversation. 


To facilitate this task, we present a new corpus named \textit{GupShup}, which contains over 6,800 Hindi-English code-switched conversations and corresponding human-annotated summaries in English as well as Hindi-English. We build this dataset by manually translating a subset of conversations and summaries from the SAMSum corpus \cite{gliwa2019samsum} from English to Hindi-English. This effort helped develop the first code-switched conversation summarization corpus containing 76,330 utterances and also provides a parallel corpus of English and Hindi-English conversations with their summaries. Following are some of the main contributions of this work:

\noindent $\bullet$ We present the first open-domain code-switched conversation summarization dataset - \textit{GupShup}, containing over 6,800 Hindi-English conversations with 76,330 utterances, and their corresponding human-annotated summaries in English and Hindi-English.

\noindent $\bullet$ We provide a thorough analysis of the dataset and performances of different state-of-the-art abstractive summarization models for the task of generating English and Hindi-English summaries from code-switched Hindi-English conversations.

\noindent $\bullet$ We also perform human evaluation of the automated summaries and present quantitative and qualitative analysis of the results.

\section{Background}
\label{background}
\begin{table*}[ht]
\centering
\scalebox{0.90}{
\begin{tabular}{|l|c|l|}
\hline
\textbf{Dataset}  & \textbf{Conversational} & \textbf{Task}\\ 
\hline
\cite{Das2014IdentifyingLA} & {\color{red}\xmark} & Language identification and POS tagging  \\ 
\cite{Barman2014CodeMA}  & {\color{red}\xmark} & POS tagging  \\ 
\cite{Chandu2015AnswerKT}  & {\color{red}\xmark} & Question Answering  \\ 
\cite{Jamatia2015PartofSpeechTF} & {\color{red}\xmark} & Language identification \\
\cite{Jamatia2016CollectingAA} & {\color{red}\xmark} & Language identification \\
\cite{Banerjee2016TheFC} & {\color{red}\xmark} & Question Answering  \\
\cite{Chakma2016CMIRAC} & {\color{red}\xmark} & Information Retrieval \\
\cite{Patro2017AllTI}  & {\color{red}\xmark} & Language identification  \\ 
\cite{Bohra2018ADO} & {\color{red}\xmark} & Hate-speech Text Classification  \\ 
\cite{Gupta2018TransliterationBT} & {\color{red}\xmark} & Question Answering  \\ 
\cite{banerjee2018dataset} & \cmark & Close domain conversation system \\ 
\cite{Chandu2018CodeMixedQA} &  {\color{red}\xmark} & Question Answering  \\ 
\cite{Patra2018SentimentAO} & {\color{red}\xmark} & Sentiment analysis \\ 
\cite{Singh2018NamedER} & {\color{red}\xmark} & Named Entity Recognition. \\
\cite{Bhat2018UniversalDP} & {\color{red}\xmark} & Dependency parsing \\ 
\cite{Bawa2018AccommodationOC}  & \cmark & Accommodation quantification \\
\cite{Khanuja2020AND} & \cmark & Natural Language Inference \\ \hline
\textbf{GupShup} (This work) & {\color{black}\cmark} & \textbf{Open domain conversation summarization} \\ \hline
\end{tabular}}
\caption{Comparison of GupShup with existing datasets on Hindi-English code-switched language tasks.}
\label{tab:literature}
\end{table*}


In the linguistics community, code-switching typically refers to the change of language or grammatical systems from one utterance to another within the same conversation \cite{gumperz1982discourse}. On the other hand, code-mixing refers to the use of linguistic units such as phrases, words, and morphemes of one language in the utterance of another language \cite{myers1997duelling,myers2002contact}. In other words, code-switching is an inter-utterance, and code-mixing is an intra-utterance phenomenon. However, in this paper, we use the term code-switching to refer to both these concepts.

Code-switching has started to gain some attention from computational linguists over the last few years \cite{barman2014code,Das2014IdentifyingLA,bali2014borrowing}, where they have been developing datasets for many interesting problems as listed in Table \ref{tab:literature}. In addition to these datasets, researchers have also started to develop objective metrics that characterize the complexity of code-switching in a given corpus \cite{gamback2016comparing,guzman2017metrics}. For a thorough review of datasets and other developments in this space, we recommend the review paper from \cite{sitaram2019survey}.

Most of the code-switched datasets typically contain individual posts or comments from social media applications like Twitter and Facebook annotated for different NLP tasks. There are very few code-switched datasets that contain complete conversations, as in back and forth utterances from multiple participants. Most notable example are: (1) Bangor Miami corpus \cite{margaret2014building}, which contains audio recordings and their corresponding transcripts of informal Spanish-English conversations between two or more speakers, (2) COMMONAMIGOS corpus \cite{ahn2020code}, which contains 587 human-computer code-mixed (Spanish and English) conversations between human users and a dialogue system, (3) DSTC2 corpus \cite{banerjee2018dataset}, which contains translations of the restaurant reservation dataset to the code-switched version using multi-lingual annotators. As shown in Table \ref{tab:literature}, there are only three datasets with Hindi-English code-switched conversations, but none of them contain summaries.

A large majority of research in abstractive summarization focuses on summarizing news articles \cite{hermann2015teaching,grusky2018newsroom,narayan2018don} and scientific papers \cite{cohan2018discourse}, mainly because of the availability of large benchmark datasets. The task of summarizing open-domain multi-party conversations has not been investigated until recently with the introduction of \textit{SamSUM} \cite{gliwa2019samsum}: a large scale English corpus of 16,000 conversations and human-annotated summaries. \cite{chen2020multi} obtained state-of-the-art results on the SamSUM corpus with their multi-view sequence-to-sequence model that encodes conversational structures by attending to topical flow and conversational stage flow in the decoder. 


To our knowledge, this work presents the first code-switched conversation summarization dataset. We also report the performances of many state-of-the-art models for generating English and Hindi-English summaries from Hindi-English code-switched conversations. In addition to the proposed code-switched conversation summarization problem, this dataset also provides an opportunity to study some other interesting research problems such as cross-lingual summarization, translation, and transliteration.

\section{Problem Statement}
\label{task}


We define a multi-party code-switched conversation $C$ as a sequence of $n$ utterances $\{u_1, u_2, ..., u_n\}$, where the $i^{th}$ utterance $u_i$ is written by $p_j$ one of the $k$ participants. We define an utterance $u_i$ as a sequence of $m$ tokens $\{x_1, x_2, ..., x_m\}$, where the tokens could either be in English or transliterated from Hindi. The goal of the code-switched conversation task is to generate a summary, a sequence of English tokens, that best captures the most salient aspects of the conversation $T$. We approach this problem as an abstractive summarization task solved using transformer-based encoder-decoder models. Since we have parallel summaries, we also consider the task of generating Hindi-English summaries (Section \ref{sec:exp}), although that is not the main focus of the paper. We analyze the predictions of different summarization models and present a brief analysis in the Appendix.


\section{Challenges}
Summarizing conversations comes with many challenges. Being informal, verbose, repetitive, sprinkled with false-starts, backchanneling, reconfirmations, hesitations, speaker interruptions, and many implicit connotations \cite{sacks1978simplest}, it is difficult for current summarization approaches to identify the most relevant and salient information from the conversations \cite{chen2020multi}. The code-switched nature of our data poses additional challenges due to the presence of multiple languages. The amount and pattern of code-switching could depend on various aspects: conversation topic, interlocutor's linguistic repertoire, the power relationship between speakers, linguistic distance, age, and the relative degree of fluency in the languages involved \cite{weninger2007speakers,muysken2000bilingual,gardner2009code}. Speakers conversant with more than two languages can also produce trilingual code-switched text, adding more complexity. In this work, we only focus on code-switching by bilingual speakers. We believe the task's challenging nature would encourage the development of better multi-lingual language models and facilitate a new direction of research in the area of summarization, leading to innovations in modeling architectures, especially for code-switched and code-mixed text.

\section{Data Collection}
\label{sec:data_collection}

One possible approach to developing a code-mixed conversation summary dataset is to write summaries for an existing code-switched conversational dataset. For example, we could start from the DSTC2 restaurant reservation dataset from \cite{banerjee2018dataset}, which is substantially large with over 50,000 utterances. However, one of the challenges with this dataset is that it focuses only on restaurant reservations; therefore, lacks linguistic diversity. Also, the conversations are between a human and a chat bot. While this is the right candidate for developing task-oriented dialogue systems, it is not suitable for open-domain code-switched conversation summarization. With this in mind, we chose a different approach of manually translating a monolingual conversation summarization corpus. As discussed earlier, the SAMSum corpus \cite{gliwa2019samsum} was clearly the best option for translation.

\begin{table*}[h]
     \centering
     \small
     \scalebox{1.0}{
     \begin{tabular}{|p{6cm}|p{6cm}|}
     \hline
 \textbf{Karen}: Hey guys! Is anyone in the office. I forgot my key... :/  & \textbf{Karen}: \textit{Hey guys! Kya koi office mein hai. Main apni chaabi bhool gayi... :/} \\
 \textbf{John}: I'll be there in 1 hour. & \textbf{John}: \textit{Main waha pe hounga in 1 hour.} \\
 \textbf{Patrick}: Oh no! I'm sorry, can't help you. I'm out of office today. & \textbf{Patrick}: \textit{Oh no! I'm sorry, main help nahi kar sakta. Maine office se bahar hu aaj.} \\
 \textbf{Mary}: Are you by the entrance? I should be there soon. & \textbf{Mary}: \textit{Kya tume entrance pe ho? I should be there soon.} \\
 \textbf{Karen}: Thanks Mary, yes, I'm here. & \textbf{Karen}: \textit{Thanks Mary, yes, main yaha hu.} \\
 \textbf{Mary}: I think I see you. 2 minutes I'm there. & \textbf{Mary}: \textit{I think main tumhe dekh sakti hu. 2 minutes, main waha hu} \\
 \textbf{Karen}: Thanks a lot! & \textbf{Karen}: \textit{Thanks a lot!} \\
 \hline
 Karen forgot the key to the office. Mary will be there soon to let her in. & \textit{Karen apni chaabi bhool gayi office ki. Mary waha pe hogi thodi der mein usko andar aane ke liye.} \\
 \hline
 \end{tabular}}
 \caption{Sample conversation and summary from SAMSum \cite{gliwa2019samsum} (first column) and the code-mixed counterpart in GupShup (second column).}
 \label{tab:sample_code_switch}
 \end{table*}

\subsection{Annotation Process} The goal of our annotation process was to build a code-switched conversation summarization corpus parallel to the SAMSum corpus. To this end, we hired eight annotators as an intern at MIDAS Lab, IIIT-Delhi for a span of 3 months. All of them were fluent in both Hindi and English. We first explained them about the concept of code-switching and provided them with a few reference examples annotated by the authors. Based on our interactions with the annotators, we observed that code-switching was an integral part of their vernacular. They also frequently used code-switching on social media and chat applications.

Each annotator was provided with a random sample of ten conversations from the SAMSum corpus. We instructed them to first go through the entire conversation and the corresponding summary in English. They were further instructed to translate the content to Hindi-English assuming it was an interaction between themselves and their friends. They performed utterance by utterance translation. The main focus was to preserve the meaning and topic of each utterance in the given context of the full conversation. Therefore, it is expected to have different length conversations in the parallel data, but the number of utterances per conversation remain the same. We asked the annotators to transcribe the resulting conversations only in Romanized text: transliterate the Hindi words as shown in Table \ref{tab:sample_code_switch}. The same process was repeated to produce the summary. Devanagari script was not used. We did not follow any strict transliteration rules during the annotation process. 

After the annotators completed annotating the initial random samples, we provided feedback in terms of format and organization of the data. Once they were comfortable with the process, we assigned them random batches of conversations, and they worked independently based on their schedules. The contributions of each annotator was mainly driven by their availability. Due to time and resource constraints, we chose to have only one translation for a given source conversation. The entire annotation process lasted for around three months, at the end of which, we translated \textit{6,831 conversations containing 76,330 utterances}, making it the first conversation summarization corpus in Hindi-English code-switched language. To our knowledge, this also makes it one of the largest parallel corpus for code-switched Hindi-English and English languages having 109,346 sentences with 48,578 of them being code-switched. 

Table \ref{tab:sample_code_switch} shows a sample conversation and summary in English and the corresponding code-switched translations. As demonstrated in this example, the annotators did minimal paraphrasing during translation and preserved the use of punctuation and emojis in the original text. This sample also demonstrates different types of code-switching patterns. For example, in the first utterance, an English term \textit{office} is inserted in the middle of a transliterated Hindi sentence. The same applies to the phrase \textit{1 hour} in the second utterance. In the fourth utterance, the speaker \textit{Mary} switches from Hindi to English in the middle of an utterance. In the last utterance, the speaker \textit{Karen} switched entirely to English though they initiated the conversation in Hindi-English.

Hindi and many other Indian languages exhibit inflection of verbs based on the gender of the speaker. For example, the sentence \textit{I read}, when said by someone of masculine gender in Hindi, would be: \textit{main padhtaa hoon}, but when said by someone of the feminine gender would be: \textit{main padhtee hoon}. The verb \textit{padh} was inflected by the gender of the speaker. During the annotation process, we did not provide explicit instructions about this inflection, but the annotators used the speaker's names or conversational context to derive the gender and used that information for translation. For example, the term \textit{hounga} in the second utterance of the conversation in Table \ref{tab:sample_conversation} is a reflection of John's perceived masculine gender. Likewise, the term \textit{sakti} in the sixth utterance is a reflection of Mary's perceived gender.

\section{Corpus Analysis}
\label{sec:data_stat}
\subsection{Conversational Diversity}
\begin{figure*}[!htb]
\centering 
\subfloat[Distribution of conversation lengths]{
    \label{fig: turn_dist}
	\includegraphics[width=0.3\textwidth]{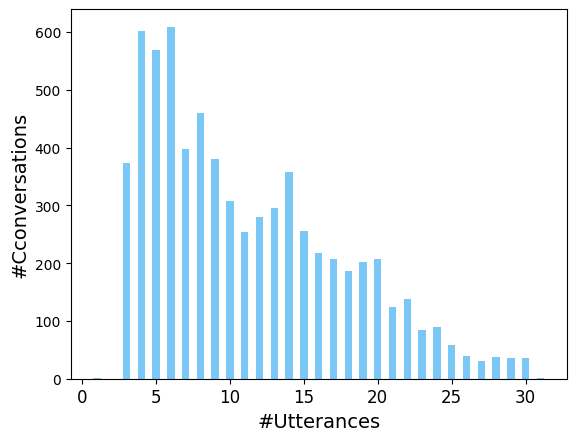}
	} 
\hspace{-0.3cm}
\subfloat[Distribution of named entities]{
    \label{fig: NE_dist}
	\includegraphics[width=0.29\textwidth]{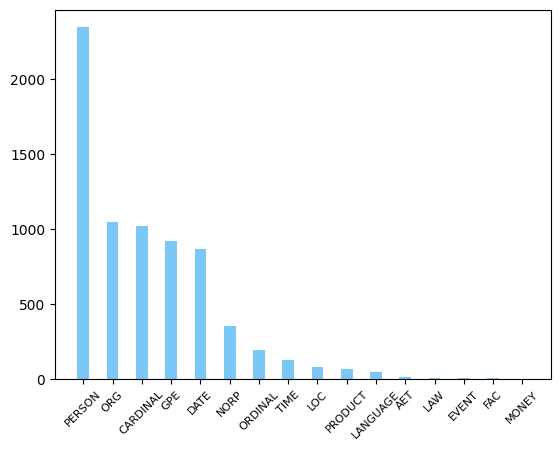}
	} 
\hspace{-0.3cm}
\subfloat[Distribution of number of speakers]{
    \label{fig: speaker_dist}
	\includegraphics[width=0.3\textwidth]{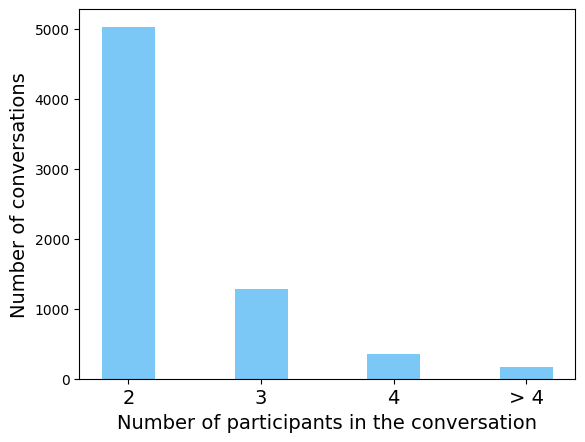}
	} 
\caption{\footnotesize \ref{fig: turn_dist} shows the distribution over number of utterances per conversation across GupShup. \ref{fig: NE_dist} shows the distribution of different named entity types across the corpus demonstrating the linguistic diversity of the utterances. \ref{fig: speaker_dist} demonstrates the distribution of number of speakers interacting in a conversation.}
\label{fig:data_stats}
\vspace{-0.3cm}
\end{figure*}
For any multi-party conversation dataset, it is important to have diversity in terms of conversation lengths: number of utterances in a conversation. Figure \ref{fig: turn_dist} shows the distribution of the number of utterances per annotated conversation. Gupshup, in total has 76,330 utterances, where the shortest conversation has 1 utterance and the longest conversation has 31 utterances. The average length of a conversation is 11.17 utterances. 

Besides length, another important diversity metric is the \textit{number of atoms (ex. words) in a conversation}, which somewhat captures the span of the conversation. If the utterances are very short, the dataset would not be useful for capturing natural human conversations. A rich conversation corpus should have diversity in terms of both conversational length and utterance length. As shown in Table \ref{tab:samsumvsgupshup}, the original English corpus SamSUM had 28.6 words per utterance, whereas code-switched GupShup has 31.1 words per utterance. Likewise, the English conversations had 19,252 unique words, but the Hindi-English version had 25,865 unique words. 

We also analyzed GupShup in terms of the \textit{number of speakers in a conversation}, which is another important aspect of multi-party conversations. A large portion of the data (73.6\% of conversations) had only two participants, 18.7\% of the conversations had 3 participants, 5.2\% had 4 participants, and the remaining conversations had more than 4 participants. 

\begin{table}[!htb]
\centering
\scalebox{0.73}{
\begin{tabular}{|l|l|l|} 
\hline
& \textbf{SamSum} & \textbf{GupShup}  \\ 
\hline
Number of utterances & 76,330 & 76,330    \\ 
\hline
Number of unique utterances & 75,696 & 75,791    \\ 
\hline
Average number of utterances per dialog & \textcolor[rgb]{0.2,0.2,0.2}{11.17}             & 11.17    \\ 
\hline
Average number of words per utterance   & \begin{tabular}[c]{@{}l@{}}28.58\\\end{tabular} & 31.12    \\ 
\hline
Average number of words per dialog     & 319.28 & 347.77   \\ 
\hline
Vocabulary Size & 19,252 & 26,865    \\
\hline
\end{tabular}}
\caption{Statistics of the subset of the English SamSum corpus compared to the code-switched Hindi-English translated version.}
\label{tab:samsumvsgupshup}
\end{table}

\subsection{Language Tagging}
Since GupShup is a code-switched corpus, it is also important to 
analyze and quantify the complexity of code-switching in the corpus. 
This requires identifying the language associated with each token in the code-switched conversations.
Since this is a parallel corpus, we used the source utterance to identify English, and that also readily identifies non-English tokens in the translated utterance. 

More precisely, for a given pair of utterances (English and Hindi-English written in Roman script), we first removed all the punctuations, emojis, and numerical tokens. We then identified all the named entities in the English utterance, and if these entities also appeared in the translated utterance, the corresponding tokens were excluded from the language tagging process. Of the remaining tokens in the Hindi-English utterance, if any of them appeared in the corresponding English utterance, we tagged them as English. 
We repeated the same process on the remaining tokens but with their lemmatized versions. This is because we observed that in some translations, the annotators used the English word from a source utterance in a different part of speech. For e.g. the phrase \textit{"a few years after we graduated"} was translated to \textit{"graduation ke kuch saal baad"}. Here graduation can be accurately tagged as English using lemmatization. Lastly, the remaining tokens in the Hindi-English utterance were tagged as Hindi. We didn’t perform lemmatization for the romanized Hindi words as there are no existing resources for doing so. Also, since there are no standard spellings for romanized Hindi words, it is possible that the high number of vocabulary is due to two very different spellings for the same word.

We observed that this algorithm captured most English tokens in the translated utterances, except for when the annotators introduced a new English token. E.g., in one of the conversations, the phrase \textit{"I have been thinking for a while"} was translated to \textit{"Main kaafi time se soch rahi hu"}, where the Hindi-English version introduced the token \textit{time}. However, we observed that this was a rare phenomenon and would not sway the analysis significantly. 

\begin{table}
\centering
\scalebox{0.75}{
\begin{tabular}{|l|l|}
\hline
Vocabulary size                                    & 26,865       \\ \hline
Code-Mixed English vocabulary                      & 11,616       \\ \hline
Romanized Hindi language vocabulary                          & 12,016       \\ \hline
Others vocabulary                                  & 1,031        \\ \hline
Total utterances                                   & 76,330       \\ \hline
Unique utterances                                  & 75,791       \\ \hline
Other utterances                                   & 1,359        \\ \hline
Code-mixed utterances                              & 43,407       \\ \hline
Romanized Hindi language utterances                     & 13,760       \\ \hline
English utterances                            & 17,804       \\ \hline
Avg. length of utterances                       & 10.07 \\ \hline
Avg. no. of code-mixed utterances per dialog & 6.35 \\ \hline
Percentage of code-mixed utterances                & 56.86\%     \\ \hline
Percentage of romanized Hindi utterances.                    & 18.02\%     \\ \hline
Percentage of English utterances                   & 23.32\%     \\ \hline
No. of code-mixed utterances with Hindi matrix     & 45,644       \\ \hline
No. of code-mixed utterances with English matrix   & 2,934        \\ \hline
No. of Hindi insertions into English Matrix     & 2,810        \\ \hline
No. of English insertions in Hindi Matrix       & 38,539       \\ \hline
No. of alternations                                & 6,853        \\ \hline
\end{tabular}}
\caption{Detailed statistics of the GupShup dataset.}
\label{tab:D_stat}
\end{table}

\subsection{Code-mixing statistics} 
Using the token-level language tagging approach described above, we further analyzed the code-switched utterances to quantify the mixing. We observed that 18.15\% (13,760) of the utterances were entirely in romanized Hindi, as in all the tokens were classified as romanized Hindi. 23.49\% (17,804) of the utterances were entirely in English. The majority of the utterances, 58.86\% (43,407), were code-mixed: had a combination of romanized Hindi and English tokens. We refer the reader to Table \ref{tab:D_stat} for more detailed code-mixing statistics of the corpus. 

One approach to quantify code-switching is through matrix language \cite{myers2002contact}, which represents the underlying language choice, therefore, driving the grammatical structure of a sentence. Determining the matrix language would ideally require sentence-level human annotations. However, since this is a costly process, we instead relied on the heuristics proposed in \cite{dhar2018enabling}. 
In particular, we define any sentence as Hindi if (a) the majority of the tokens are romanized Hindi, (b) we detect the use of any romanized Hindi verbs, or (c) we detect the use of any romanized Hindi bi-grams. 
Per this definition, the matrix language of 45,644 sentences was Hindi, and 2,934 sentences were in English. Please note that these numbers are specified at sentence-level and not utterance-level: they correspond only to the 43,407 code-mixed utterances, where each utterance can have multiple sentences.  Based on the matrix language,  one can further quantify the complexity of code-switching in the corpus using the metrics proposed in \cite{gamback2016comparing}. 

\begin{table} [!htbp]
\centering
\scalebox{1.0}{
\begin{tabular}{|c|c|c|} 
\hline
 \textbf{I-index} & $\bf C_{c}$ & $\bf C_{avg}$  \\ 
\hline
 0.14      &   63.25 &   13.57  \\ 
\hline
\end{tabular}}
\caption{Code mixing metrics for the code-switched Hindi-English conversations (GupSHup)}
\label{tab:C_stats}
\end{table} 

\begin{table*}
    \centering
    \scalebox{1.0}{
    \begin{tabular}{ |l|c|c|c|c|c|c|c| } 
    \hline
    Model & R1 & R2 & RL & BLEURT & BERTScore & BLEU  & METEOR\\
    \hline
    mBART & \bf{43.14}  & \bf{16.83} & 33.87       & -0.46      & 0.9   & 9.96              & 26.74      \\
    Multi-view & 41.21   & 16.16      & \bf{39.80}  & \bf{-0.43} & 0.9   & \bf{11.45}        & 28.75      \\
    PEGASUS & 41.6      & 15.72      & 32.91       & -0.44      & 0.9   & 8.91              & 25.55      \\
    T5 MTL  & 40.84     & 15.50      & 30.38       & -0.47      & 0.9   & 11.05             & \bf{30.8}  \\
    T5      & 37.52     & 12.60      & 27.55       & -0.56      & 0.89  & 8.12              & 26.92      \\
    BART    & 39.75     & 14.09      & 31.52       & -0.52      & 0.9   & 6.92              & 23.73      \\
    GPT2    & 13.52     & 2.59       & 10.5        & -1.03      & 0.84  & 2.21              & 12.05      \\
    
    \hline
    \end{tabular}}
    \caption{Performances on automatic metrics for generating English summaries from Hindi-English conversations.}
    \label{tab:results}
\end{table*}

\begin{table*}[!h]
    \centering
    \scalebox{1.0}{
    \begin{tabular}{|l|c|c|c|c|c|c|c|}
    \hline
        Model & R1 & R2  & RL & BLEURT & BERTScore & BLEU & METEOR \\ \hline
        Multi-view & \bf{50.65} & 25.04 & 40.13 & -0.30
 & 0.92 & \bf{18.34} & \bf{39.04} \\ 
        PEGASUS & 50.53 & \bf{25.77} & \bf{41.94} & \bf{-0.28} & 0.92 & 17.47 & 36.64 \\ 
        mBART & 49.69 & 24.36 & 40.40 & -0.35
 & 0.91 & 14.37 & 32.44 \\ 
        T5  & 45.62 & 21.65 & 35.25 & -0.46 & 0.91 & 14.95 & 38.24 \\ 
        BART & 46.47 & 21.79 & 37.74 & -0.39 & 0.91 & 14.37 & 32.44 \\ 
        GPT2 & 15.78 & 5.42 & 13.58 & -0.98 & 0.84 & 2.78 & 14.75 \\
        \hline
    \end{tabular}}
    \caption{Performances on automatic metrics for generating English summaries from English conversations.}
\label{tab:e2e}
\end{table*}

\begin{table*}[h!]
    \centering
    \scalebox{1.0}{
    \begin{tabular}{ |l|c|c|c|c| } 
    \hline
    Model & Coherence & Consistency & Fluency & Relevance \\
    \hline
    mBART     & 0.81       & 0.63      & 0.85       & \bf{0.65} \\
    Multi-view & 0.83       & \bf{0.65} & 0.86       & \bf{0.65}   \\
    PEGASUS   & \bf{0.84}  & 0.58      & \bf{0.87}  & 0.60   \\
    T5 MTL    & 0.70       & 0.55      & 0.77       & 0.55   \\
    T5        & 0.67       & 0.532     & 0.77       & 0.54   \\
    BART      & 0.73       & 0.56      & 0.74       & 0.55      \\
    GPT2      & 0.37       & 0.36      & 0.52       & 0.33   \\
    
    \hline
    \end{tabular}}
    \caption{Human evaluation of the generated summaries (English) from code-switched Hindi-English conversations.}
    \label{tab:humaneval}
\end{table*}

The utterance level code-switching complexity $C_u$ depends on the number of foreign language tokens, language switch points, and the total number of language dependent tokens in the utterance. $C_u$ will be 0 if the utterance is monolingual or contains only language independent tokens. The higher the $C_u$ metric, the more complex is the utterance's code-mixing.

Using $C_u$, we can calculate the code-switching complexity at the corpus level. $C_{avg}$ is the average of $C_u$ values of all the utterances of the corpus. $C_c$ is a slight modification of $C_{avg}$, where \cite{banerjee2018dataset} remove the assumption that the language with maximum tokens is the matrix language. We also calculate I-index\cite{Guzmn2016SimpleTF} which can be calculated simply by computing the number of switch points in the corpus. It quantifies the integration of different languages in the corpus.
On our corpus, we estimated that $C_c$ was 63.25 and $C_{avg}$ was 13.57 (Table \ref{tab:C_stats}). 

We further analyzed the data in terms of named entities mentioned during conversations. We define named entity as one identified both in English, using Spacy's named entity detection algorithm, and the translated Hindi-English utterance based on token matching with the entity identified in the English utterance. Figure \ref{fig: NE_dist} shows a distribution of different types of entities identified in the corpus. The most frequently mentioned entity type is Person, which is expected considering the nature of the conversational data where references to other participants are very common.

\section{Empirical Benchmarks}
\label{sec:exp}
The main goal of our experimental work is to validate some of the state-of-the-art abstractive summarization models on the new dataset. We expect these results to serve as empirical benchmarks for future researchers. To this end, we experimented with various encoder-decoder transformer models to generate English summaries from code-switched conversations. Most of these transformer models were pre-trained on English corpora, except for mBART, which was trained on multilingual data.

\begin{table*}[h!]
    \centering
    \scalebox{1.0}{
    \begin{tabular}{ |l|c|c|c|c|c|c|c| } 
    \hline
    Model & R1 & R2 & RL & BLEURT & BERTSCORE & BLEU & METEOR  \\
    \hline
    Coherence     & 0.86	& 0.82	&0.89	&0.93	&0.8	&0.57	&0.18\\
    Consistency   & 0.86	& 0.93	&1	    &0.93	&0.8	&0.71	&0.32\\
    Fluency       & 0.86	& 0.82	& 0.79	&0.93	&0.67	&0.75	&0.46\\
    Relevance     & 0.89	& 0.96	&0.96	&0.96	&0.8	&0.86	&0.5\\
    Overall       & 0.89	&\bf 0.96	&\bf 0.96	&\bf 0.96	&0.8	&0.86	&0.5\\	
    
    \hline
    \end{tabular}}
    \caption{Correlation between human judgement and automatic metrics for English summaries generated from code-switched Hindi-English conversations.}
    \label{tab:metric_correlation}
\end{table*}

\begin{table*}[ht]
\centering
\scalebox{1.0}{
\begin{tabular}{|l|c|c|c|c|c|c|c|}
\hline
Models & R1 & R2 & R3 & BLEURT & BERTScore & BLEU & METEOR \\ \hline
mBART & 19.98 & 2.89 & 16.7 & -0.88 & 0.83 & 1.60 & 10.42 \\ \hline
PEGASUS & 35.69 & 11.01 & 28.78 & -0.72 & 0.86 & \textbf{6.16} & 20.91 \\ \hline
Multi-view & 21.92 & 4.55 & 18.16 & -0.83 & 0.85 & 2.27 & 9.94 \\ \hline
BART & \textbf{36.28} & \textbf{11.45} & \textbf{28.92} & \textbf{-0.70} & \textbf{0.87} & 5.96 & \textbf{21.82} \\ \hline
T5 & 31.85 & 7.90 & 24.13 & -0.72 & 0.86 & 5.51 & 20.6 \\ \hline
\end{tabular}}
\caption{Performances on automatic metrics for generating Hindi-English summaries from Hindi-English conversations.}
\end{table*}

\subsection{Experimental Settings}
For this benchmarking effort, we experimented with the following models: GPT-2, BART, PEGASUS, T5, multitask T5, and mBART. We also applied the multi-view seq2seq model \cite{chen2020multi}, which achieved the state-of-the-art performance on the SamSum corpus. We trained all the models for 3 epochs, with evaluation on validation set after each epoch. We used 5,831/500/500 conversations as our train, dev and test splits, respectively. For mBART, a smaller scale mBART with 12-layer encoder and 6-layer decoder was implemented due to lack of computing resources. We've also tried two different training approaches of T5, a pre-train-then-fine-tune approach as well as multi-task training approach (T5 MTL), in which we train for both the tasks of summarization and translation. All models were trained using the Huggingface's transformer library, on Google colab GPU enabled platform. All the models were trained using the basic colab account available for everyone.



To fully characterize the dataset, we've sought both automatic and human evaluation. For Automatic evaluation, we considered ROUGE (R1, R2, RL) \cite{lin-2004-rouge}, BLEURT \cite{sellam2020bleurt}, BERT-score \cite{bert-score}, BLEU \cite{Papineni2002BleuAM}, and METEOR \cite{banerjee-lavie-2005-meteor}. Then, these metrics were compared against human evaluation for correlation as well as interpretation with respect to human judgement. The human evaluation was conducted for 4 possible metrics: \textit{coherence}, \textit{consistency}, \textit{fluency}, and \textit{relevance}.

\subsection{Results}
The final results of the experiments evaluated on automatic metrics are shown in Table \ref{tab:results}. On the automatic metrics, mBART and Multi-view performed best on recall based summarization metrics, R1, R2, and RL. Since mBART is the only model trained on multiple languages, it may have learned the location of English and Hindi switching better than other models exclusively trained on English. On precision based BLEU, BLEURT, and RL, as well as in human evaluation, Multi-view performed the best overall. Though it was pre-trained on English corpora, its decoding strategy that focuses on various aspects, or ``views" on the conversation proved to be effective in understanding the conversations. Perhaps, we may investigate incorporating additional Hindi corpus to pre-training Multi-view seq2seq model as a next step.

The T5 MTL model performed significantly better than the T5 model on all the automatic metrics as well as human evaluation. We applied the Wilcoxon-signed rank test, using which we calculated the p-value to be 7.1e-16 at a confidence level of 5\%, indicating that there is a statistically significant difference between the ROUGE-1 scores produced by both the models on 500 test examples. This also indicates that due to the multi-lingual and parallel nature of the dataset, the summarization task would benefit from the translation task. It would be interesting to try out other models in a multi-task learning setup and investigate more in the future. 

For a better understanding of the complexity of the task at hand, i.e. summarizing conversations, we also trained these pre-trained encoder-decoder models for summarizing the parallel English conversations into their respective English abstractive summaries. Table \ref{tab:e2e} shows the results for this summarization task.
We observe that the performance of all the models increases by a good margin. This highlights that code-switching could be one of the factors impacting the performance of the models for the task of summarization. Multi-view seq2seq seems to be the best performing model in this case as well, compared to summarizing code-switched Hindi-English conversations to English summaries. Pegasus also has a comparable performance. It is surprising to see that mBART is still competent with the other models when used for monolingual data alone. The gap between the scores of BART and mBART also seems to be much lesser when only English data is used for summarization. It is unclear if the superior performance of mBART is due to its multilinguality, or that it is pre-trained on more English data when compared to BART. The Hindi text in the code-switched conversations are in Roman script which enabled us to use monolingual models, which appear to be good contextualizers in this case. We would further like to explore using Devanagari script for the Hindi text in the code-switched conversations for the multilingual model, to find out if the Devanagari helps the model deduce better meaning of the text, given mBART is pre-trained on Hindi written in Devanagri script.

We also generated Hindi-English summaries by training the models that performed well in the generation of English summaries. Except BART, none of the models performed as well as they performed while generating the English summaries showing a significant drop in their performances. What is more surprising is that the mBART, which is the only multilingual model in the list, seems to perform the worst when it comes to generating code-switched summaries. Given it is a multilingual model, it should have been easier for mBART to decode in a language other than English as it is pre-trained on 25 languages and has shown good performance on cross-lingual transfer learning. 
Overall drop in the performance for all the models was expected as the current state-of-the-art models might not be capable of understanding code-switched Hindi-English text as they are not trained on such data. This is certainly a direction that needs a serious effort  in the near future if we want our models to understand and generate code-switched language.

\subsection{Human Evaluation}
To qualify the comparison and results, we've conducted human evaluation on a randomly selected 100 summaries from our test set as shown in Table \ref{tab:humaneval}. Following \cite{Fabbri2020SummEvalRS}, we considered 4 metrics: \textit{consistency}, \textit{coherence}, \textit{fluency}, and \textit{relevance}.  These were measured on a scale of 1 to 5, with 1 as poor and 5 as perfect. The meaning of these metrics are briefly summarized. (a) \textit{Coherence}: the collective quality of overall sentences. Measures how well the summaries are organized and structured, (b) \textit{Consistency}: measures how well the factual information was transferred, (c) \textit{Fluency}: the grammatical quality and naturalness of individual sentences, (d) \textit{Relevance}: the selectivity of summary information by importance. The summary should only contain significant information from the input.





On human evaluation, Multi-view, mBART, and PEGASUS achieved comparably good performances, with Multi-view achieving the best performance. Intuitively, mBART should be the best at identifying the transition points between English and Hindi as mBART has been trained on 
multilingual corpora, which means that it may recognize non-English words. Similarly, PEGASUS, which is trained for summarization on masked corpus, may recognize romanized Hindi phrases as a type of ``gaps" or masked-tokens embedded in English corpus. On the other hand, Multi-view model that focuses on various aspects of the conversation beyond sentence level would produce structurally sound and meaningful summaries. In terms of human evaluation, Multi-view and PEGASUS seems to have achieved best structural understanding of the dialog, reflected in high performance on the coherence metric. Moreover, mBART and Multi-view have fared well with informational significance metrics such as consistency and relevance.

Given the plethora of available automatic metrics, we have established connection with the human evaluation metrics as Spearman correlation, calculated in Table \ref{tab:metric_correlation}. The Overall correlation is calculated as the correlation between the automatic metrics and the sum of all the four human evaluation metrics (consistency, coherence, fluency, and relevance). Recall based summarization metrics and BLEURT were best correlated with human judgements. Leveraging paraphrases and sentence pairs for pretraining BERT and fine-tuned on human judgement scores, BLEURT proved to be a robust metric for this task. Perhaps as a next step, BLEURT metric could be fine-tuned on this dataset to serve as a code-switching specific metric.

\section{Conclusion and Future Work}
\label{sec:conclusion}
In this work, we presented the first code-switched conversation summarization dataset - \textit{GupShup}, having 6,831 multi-party open-domain conversations with their corresponding summaries in English and Hindi-English. We conducted a thorough evaluation of the  state-of-the-art neural models for the task of generating English summaries of Hindi-English conversations using automatic metrics as well as human evaluation. Multi-lingual mBART and multi-view seq2seq models performed the best in generating English summaries, whereas the same models showed poorer performance in generating Hindi-English summaries indicating their inability to process Hindi-English code-switched text. Multi-task learning setup showed promise and we would like to look more into it in the near future. We are also looking forward to using our dataset for the purposes of translation and data augmentation, using the 74,798 parallell Hindi-English and English sentences, derived from the 6,831 Hindi-English and English conversation pairs. 
We believe that our dataset along with its challenging task will prove to be a new benchmark for developing sophisticated NLP models with capabilities of understanding code-switched text.

\bibliography{anthology,custom}
\bibliographystyle{acl_natbib}

\appendix
\section{Appendix}
\label{sec:appendix}


\section{Qualitative Analysis of Summaries Generated by Each Model}

For each model we analysed the top 10 and the bottom 10 summaries generated in terms of  ROUGE 1. The following sections present the overall observation for each model. 

\subsection{mBART}

\subsubsection{Top 10}
\begin{itemize}
    \item The highest scoring summaries have a ROUGE 1 score of 1.0.
    \item All of the top 10 summaries belong to very small conversations with an average of just 5 utterances in the dialogue. We still notice a few cases where wrong speakers are associated with wrong actions/events.
    
Example:  \\
Eric: Wya? \\
Eve: College Green \\
Eric: \textit{wahi reh}, I'll pick you up \\
Eve: <3 \\
\textbf{Gold summary}: \textit{Eric will pick Eve from College Green.} \\
\textbf{Predicted summary}: Eve will pick Eric up from college Green.

\end{itemize}

\subsubsection{Bottom 10}

\begin{itemize}
    \item Bottom 10 summaries belonged to conversations with varying lengths. Some were very long and some very short. The level of code-mixing varied hugely as well. We will look into more concrete factors that make these conversations inherently tougher for mBART. 
\end{itemize}

\subsection{PEGASUS}
\subsubsection{Top 10}
\begin{itemize}
    \item The highest scoring summaries have a ROUGE 1 score of 1.0. All the summaries in top 10 belong to conversations that are very small. PEGASUS also produces a few summaries where wrong speakers are associated with wrong actions/events, for the same examples as that of mBART. 
 
\end{itemize}
\subsubsection{Bottom 10}

\begin{itemize}
\item Bottom 10 summaries belonged to conversations with varying lengths. Some were very long and some very short. The level of code-mixing varied hugely as well. We will look into more concrete factors that make these conversations inherently tougher for PEGASUS. 
\end{itemize}

\subsection{BART}

\subsubsection{Top 10} 
\begin{itemize}
    \item The highest scoring summaries have a ROUGE 1 score of 1.0. All the summaries in top 10 belong to conversations that are very small. Like mBART and PEGASUS, BART also produces a few summaries where wrong speakers are associated with wrong actions/events.
\end{itemize}

\subsubsection{Bottom 10}
\begin{itemize} 
\item Bottom 10 summaries belonged to conversations with varying lengths. Some were very long and some very short. The level of code-mixing varied hugely as well. We will look into more concrete factors that make these conversations inherently tougher for BART.  
\end{itemize}

\subsection{Multiview Seq2Seq}
\subsubsection{Top 10}
\begin{itemize}
    \item The highest scoring summaries have a ROUGE 1 score of 1.0. All the summaries in top 10 belong to conversations that are very small. Like mBART, PEGASUS, and BART, Multiview Seq2Seq also produces a few summaries where wrong speakers are associated with wrong actions/events.
\end{itemize}

\subsubsection{Bottom 10}
\begin{itemize}
    \item Bottom 10 summaries belonged to conversations with varying lengths. Some were very long and some very short. The level of code-mixing varied hugely as well. We will look into more concrete factors that make these conversations inherently tougher for Multiview Seq2Seq. 
\end{itemize}

\subsection{T5}
\subsubsection{Top 10}
\begin{itemize}
    \item The highest scoring summaries have a ROUGE 1 score of 0.75. Interestingly, half of the top 10 scoring summaries belong to long conversations, with an average of 10-15 utterances. This observation is different from what is noticed in mBART, PEGASUS, BART, and Multiview Seq2Seq. The rest of the summaries belong to small conversations. Like mBART, PEGASUS, BART, and Multiview Seq2Seq, T5 also produces a few summaries where wrong speakers are associated with wrong actions/events. This error seems to be consistent throughout the models. 
\end{itemize}

The best scoring summary of T5 belongs to a very long conversation. Following is the particular example: \\ 
\\
Jamal: <file\_photo> \\
Terry: Taj Mahal! \\
Maria: Yes, we visited it today with Jamal \\
Ken: \textit{yeh bahut sundar} mosque \textit{ hai}! \\
Maria: it's not a mosque! \\
Ken: what? \\
Maria: it's a mausoleum \\
Ken: \textit{mujhe hamesha lagta tha} it's a mosque \\
Jamal: \textit{bahut logo ko lagta hai} \\
Maria: it is a mausoleum that an emperor commissioned for his favourite wife \\
Maria: \textit{shayad uska naam Mumtaz Mahal tha} \\
Jamal: correct! :D what a good pupil! \\ 
Maria: haha, because it's such a romantic story \\
Maria: 20000\textit{ logo ne Taj Mahal banaya}, it's so monumental \\
Ken:\textit{ iss naam kaa kya matlab hai?} \\
Maria: Taj is a short version of Mumtaz \\
Maria: and Mumtaz Mahal means "Crown of the Palace" \\
Ken: wow \\
Maria: Jamal was an amazing guide today \\
Ken: I wish I was there with you \\

\textbf{Gold summary}: \textit{Maria and Jamal visited Taj Mahal today. It's a mausoleum that an emperor commissioned for his wife Mumtaz Mahal.}

\textbf{Predicted summary}: Maria and Jamal visited the Taj Mahal today. The mausoleum was commissioned by an emperor for his favourite wife, Mumtaz.

\subsubsection{Bottom 10}
\begin{itemize}
    \item Bottom 10 summaries belonged to conversations with varying lengths. Some were very long and some very short. The level of code-mixing varied hugely as well. We will look into more concrete factors that make these conversations inherently tougher for T5. 
\end{itemize}

\section{Few Interesting Examples}

\begin{itemize}
   
\item This conversation has a lot of content in English except the word “buy”, because of which all the models produce wrong summaries except for mBART, which is also our only multilingual model.   \\
\\

Nick: Opinions required! Gas \textit{ya} induction hob? \\
Ben: \textit{Bahut} time \textit{se ek} induction hob use \textit{kar raha huu}, \textit{mai} convinced\textit{ nahi huu}.. \\
Ruth: induction-\textit{ bahut} sleek \textit{hai jadi garam ho jaata hai!} \\
Ben: but \textit{voh} constant temperature maintain \textit{nahi karta! Kya saare} induction \textit{aise hee hai yaa maine purana waala liya tha}? \\
Ruth: they pulse \textit{ agar} proper pans \textit{nahi} use \textit{karte hum} \\
Ben: proper \textit{se} \textit{tumhara kya} \textit{matlab? Kya tumhara matlab} better \textit{aur bhari?} \\
Ruth: yeah, simply suitable \\
Ben: and \textit{mujhe lagta hai mujhe usse chalana seekhna padega..} \\
Ruth: yeah,\textit{ yeh } gas \textit{se alag hai} \\
Christian: gas, \textit{bina kisi saval ke- aur koi cheez tumhe}  control \textit{nahi deti!} \\
Nick: \textit{Mai usme} interested \textit{huu jisme mujhe mere hissab se} consistent heat \textit{ mile} \\
Mary: with induction \textit{tumhe usse shuru aur band karna padega} to regulate temperature.. \\
Kate: induction- yes, gas- no\textit{ kyuki paani ubalne mein sadiyaan lag jaati hai!} \\
Tim: \textit{tum jaante ho naa tum ek} electric kettle use \textit{kar sakte ho?} \\
Kate: haha!\textit{ Yeh} funny \textit{nahi hai!} \\
Kate: \textit{yeh saaf karne mein bhi aasan hai.} \\
Harry:\textit{ mai } induction\textit{ lunga kyuki voh khaana banane ke baad bhi usse garam rakhta hai} \\
Tom: Induction! 100\% \\
Susan: \textit{humara} induction \textit{bahut ganda tha} I think it's common! \\
Emma: induction \textit{ke liye ek aur} vote!  \\
Ruth: All chefs seem to say gas! \\
Tom: \textit{mai} gas \textit{se zyaada} induction hobs \textit{bechta huu! yeh} popular\textit{ hota jaa raha hai aur mai dekh ssakta huu kyu!} \\
Emma:\textit{ humne humara waala} John Lewis outlet \textit{se liya so} it was ex display \textit{aur isliye bahut sasta tha!} \\
Nick:\textit{ tumhari raaye ke liye } thanks! Great talk! \textit{Mujhe lagta hai mai} Induction lunga. \\

\textbf{Gold summary}:\textit{Nick decided to buy an induction hob.} \\
\textbf{mBART:}  Nick wants to buy an induction hob. He thinks induction is better than gas. \\
\textbf{Pegasus: } Ben, Ruth, Tom, Emma and Kate are debating whether or not they should use gas in their induction. \\
\textbf{T5:}  Ruth and Christian are interested in induction. They use a heat band to regulate the temperature. Kate is interested in the idea of induction, but she doesn't like it. \\
\textbf{T5 Multitask setup:} Ben and Ruth use induction hobs, but they don't maintain a constant temperature. Nick and Kate are interested in induction. Harry uses induction, but he doesn't like it. \\
\textbf{BART:} Ruth, Nick, Mary, Tom, Emma and Emma are interested in induction hobs. \\
\textbf{Multiview Seq2Seq}: Ruth, Ben, Tom, Tom and Ben are not convinced about the induction hob. They want to know if it is better to use gas or induction. \\

\item This conversation is present in top 10 in both PEGASUS and BART, but BART confuses between the entities:

Emir: Etna \textit{ki} financial statement\textit{ bhej sakti ho?} \\
Britta: Sure, \textit{konsa saal?} \\
Emir: 2017 \\
Britta: Ok \\
Emir: English \textit{mei} please \\

\textbf{Gold summary:} \textit{Britta will send Emir Etna's 2017 financial statement in English.} \\
\textbf{PEGASUS:} Britta will send Emir the financial statement from Etna in 2017 in English. \\
\textbf{BART:} Emir will send Britta the financial statement for 2017 in English. \\

Apart from this, PEGASUS, BART, and mBART have 7 common examples in their top 20. PEGASUS and mBART have 9 examples that are common in top 20. BART and PEGASUS have 9 examples common as well. BART and mBART have 11 examples common in their top 20. 

Overall, similar pattern is observed in top 10 examples for all the models(mBART, PEGASUS, BART, and Multiview seq2seq) except for T5. T5 is the only model that has high scoring summaries for very long conversations. 

\end{itemize}

\begin{table*}[!h]
    \centering
    \begin{tabular}{ |l|c|c|c|c|c| } 
    \hline
    Model & Learning Rate & Optimizer & No. Epochs & Batch Size & No. Beams\\
    \hline
    GPT2 & 3e-4 & Adam & 3 & 1 & 4\\
    BART & 3e-5 & Adam & 3 & 1 & 4\\
    PEGASUS & 5e-4  & Adam & 3 & 1 & 4\\
    T5 MLT & 3e-5 & Adam & 3 & 1 & 8\\
    T5 & 3e-5 & Adam & 3 & 1 & 1\\
    mBART & 3e-5 & Adam & 10 & 1 & 1\\
    Multiview & 3e-4 & Adam & 3 & 32 & 4\\
    
    \hline
    \end{tabular}
    \caption{Training parameters for the different summarization models.}
    \label{tab:train}
\end{table*}

\begin{table*}[]
\centering
\begin{tabular}{|l|l|}
\hline
Conversation  \#1   & \begin{tabular}[c]{@{}l@{}}Peter: Kya mai tumhari car borrow kar sakta hu? \\ 
Hugh: sure \\ 
Hugh: but tumhari car ke saath kya hua? \\ 
Peter: pata nahi \\ 
Peter: aur time nahi hai check karne ka \\ 
Peter: mai wese bhi late hu! \\ 
Hugh: ok, ok, aake lele\end{tabular} \\ \hline

PEGASUS generated summary \#1 & Peter will borrow Hugh's car. \\ \hline
Gold summary \#1 & Peter will borrow Hugh's car.\\ \hline

Conversation \#2  & \begin{tabular}[c]{@{}l@{}}
Monica: Kaha ho tum?\\ 
Monica: Mai tumhe dekh ni pa rhi.\\ 
Lexie: Bs yahi hu.\\ 
Monica: OK, waiting.
\end{tabular}                                                   \\ \hline

PEGASUS generated summary \#2 & Monica is waiting for Lexie. \\ \hline
Gold summary \#2 & Monica is waiting for Lexie. \\ \hline

Conversation \#3 & \begin{tabular}[c]{@{}l@{}}

Agatha: meri book read karna khatam kar diya? \\ 
Johnny: abhi nahi \\ 
Agatha: I see\end{tabular} \\ \hline
PEGASUS generated summary \#3 & Johnny hasn't read Agatha's book yet. \\ \hline
Conversation \#3 & Johnny hasn't finished reading Agatha's book yet. \\ \hline
\end{tabular}
\caption{Top 3 summaries generated by PEGASUS.}
\end{table*}

\begin{table*}[]
\centering
\begin{tabular}{|l|l|}
\hline
Conversation  \#1    & \begin{tabular}[c]{@{}l@{}}Scott: Hume kaha milna chahiye? \\ John: at Oculus? \\ Scott: ok! 7.30 baje \\ John: yup!\end{tabular}                                    
\\ \hline
mBART generated summary \#1 & Scott and John will meet at Oculus at 7.30.                                                        \\ \hline
Gold summary \#1 & Scott and John will meet at 7.30 at Oculus.   \\ \hline
Conversation \#2  & \begin{tabular}[c]{@{}l@{}}Ralph: Tum abhi bhi yahi kahi ho? \\ Mary: Bathroom! \\ Ralph: Oh! TMI! \\ Mary: ek sec mein aa raha huu… \\ Ralph: jitna time chaiye utna lo...\end{tabular}                                                \\ \hline
mBART generated summary \#2 & Mary is in the bathroom.          \\ \hline
Gold summary \#2 & Mary is in the bathroom.                      \\ \hline
Conversation \#3 & \begin{tabular}[c]{@{}l@{}}Peter: Kya mai tumhari car borrow kar sakta hu? \\ Hugh: sure \\ Hugh: but tumhari car ke saath kya hua? \\ Peter: pata nahi \\ Peter: aur time nahi hai check karne ka \\ Peter: mai wese bhi late hu! \\ Hugh: ok, ok, aake lele\end{tabular} \\ \hline
mBART generated summary \#3 & Peter will borrow Hugh's car.       \\ \hline
Gold summary \#3   & Peter will borrow Hugh's car.              \\ \hline
\end{tabular}
\caption{Top 3 summaries generated by mBART.}
\end{table*}

\begin{table*}[]
\centering
\begin{tabular}{|l|l|}
\hline
Conversation  \#1               & \begin{tabular}[c]{@{}l@{}}Scott: Hume kaha milna chahiye? \\ John: at Oculus? \\ Scott: ok! 7.30 baje \\ John: yup!\end{tabular}  \\ \hline
Multiview generated summary \#1 & John and Scott will meet at Oculus at 7.30.    \\ \hline
Gold summary \#1   & Scott and John will meet at 7.30 at Oculus.   \\ \hline
Conversation \#2                & \begin{tabular}[c]{@{}l@{}}Jude: Tumhara wallet kaha hai mujhe kahi nahi mil raha \\ Faith: maine apne saath le liya tha \\ Jude: kyu? I need your credit card to pay the bills\end{tabular} \\ \hline
Multiview generated summary \#2 & Jude needs Faith's credit card to pay the bills.\\ \hline
Gold summary \#2   & Jude needs Faith's credit card to pay the bills. \\ \hline
Conversation \#3                & \begin{tabular}[c]{@{}l@{}}Emir: Etna ki financial statement bhej sakti ho? \\ Britta: Sure, konsa saal? \\ Emir: 2017 \\ Britta: Ok \\ Emir: English mei please\end{tabular} \\ \hline
Mutliview generated summary \#3 & Emir will send Britta Etna's financial statement in English. \\ \hline
Gold summary \#3  & Britta will send Emir Etna's 2017 financial statement in English.  \\ \hline
\end{tabular}
\caption{Top 3 summaries generated by Multi-view seq2seq model.}
\end{table*}

\end{document}